\documentclass[5p, times]{elsarticle}

\makeatletter
\def\ps@pprintTitle{%
   \let\@oddhead\@empty
   \let\@evenhead\@empty
   \let\@oddfoot\@empty
   \let\@evenfoot\@oddfoot
}
\makeatother

\usepackage{bm}
\usepackage{float}
\usepackage[svgnames,dvipsnames,table]{xcolor}
\usepackage[colorlinks]{hyperref}
\usepackage{amsmath}
\usepackage{amssymb}
\usepackage{adjustbox}
\usepackage{multirow}
\usepackage{tikz}
\usepackage{pgfplots}
\usepgfplotslibrary{patchplots}
\usetikzlibrary{trees,shapes,snakes,calc,matrix,positioning,fit,arrows,patterns,backgrounds,quotes,arrows.meta}
\usepackage{siunitx}
\usepackage{multicol}
\usepackage{subcaption} 
\usetikzlibrary{shapes.geometric}
\usetikzlibrary{shapes.arrows}
\usepgfplotslibrary{fillbetween}
\usepackage{array}
\usepackage{tabularx}
\usepackage{lipsum} 
 \usepackage{collcell}
\usepackage{hhline}
\usetikzlibrary{spy, calc}
\usepackage{array}
\usepackage{tabularx}
\usepackage{tabulary}
\usepackage{xspace}

\usepackage{bbding}
\usepackage{pifont}
\usepackage{diagbox}
\usepackage{booktabs}
\usepackage{makecell}
\usetikzlibrary{fit}
\usepackage{rotating}
\usepgfplotslibrary{colorbrewer}
\usepackage{fontawesome}
\usepackage{lscape}
\usepackage[para,online,flushleft]{threeparttable}
\usepackage{longtable}
\usepackage{tablefootnote}
\usepackage{makecell}
\usepackage{breakcites}
\usepackage{pifont}
\usepackage{hyphenat}
\begin{document}
\begin{frontmatter}

\title{DL-CapsNet: A Deep and Light Capsule Network}


\author[1]{Pouya Shiri}
\ead{pouyashiri@uvic.ca}
\author[1]{Amirali Baniasadi}
\ead{amiralib@uvic.ca}

\address[1]{Department of Electrical and Computer Engineering, University of Victoria, 3800 Finnerty Rd, Victoria, BC, V8P 5J2 Canada}

\begin{abstract}
Capsule Network (CapsNet) is among the promising classifiers and a possible successor of the classifiers built based on Convolutional Neural Network (CNN). CapsNet is more accurate than CNNs in detecting images with overlapping categories and those with applied affine transformations. 
 
In this work, we propose a deep variant of CapsNet consisting of several capsule layers. In addition, we design the Capsule Summarization layer to reduce the complexity by reducing the number of parameters. 

DL-CapsNet, while being highly accurate, employs a small number of parameters and delivers faster training and inference. DL-CapsNet can process complex datasets with a high number of categories.
\end{abstract}


\end{frontmatter}


\section{Introduction}
Sabour et al. introduced Capsule Network (CapsNet) \cite{Sabour2017} as the new generation of classifiers with several advantages over traditional Convolutional Neural Networks (CNNs). CapsNet is more robust to applying affine transformations and detects images with overlapping categories easier than CNNs.  CapsNet offers competitive accuracy showing promising results on small-scale datasets such as MNIST \cite{LECUN} and Fashion-MNIST \cite{Xiao2017}. On more complex datasets such as CIFAR-10 and CIFAR-100 \cite{Krizhevsky2009}, however, the results are not as competitive. There have been several works aiming at facilitating supporting datasets with a high number of categories.

\par
The basic computational unit of CapsNet is referred to as a capsule (a vector of neurons). CapsNet consists of a simple feature extractor including two convolutional layers. The extracted features are then reshaped to vectors. These vectors are multiplied by multiple matrices to produce the first level of capsules referred to as Primary Capsules (PCs). The next layer of capsules (output capsules) are generated out of PCs using an iterative algorithm called Dynamic Routing (DR).
In DR, all input capsules contribute to all output capsules but with different weights. The output capsules are used for the classification. CapsNet employs a simple decoder consisting of Fully-Connected (FC) layers to regularize training by adding the reconstruction term to the loss function.
\par

We propose a deep network to add support for more complex datasets. Making networks deeper (stacking up layers) results in high generalization, and hence is common. However, deeper networks have higher number of parameters (trainable weights). This is critical as it affects the computation cost and the resources required. Therefore, we take further measures to reduce complexity by developing a mechanism to reduce the number of capsules. This is achieved by replacing multiple capsules with only a few using a summarization mechanism.

\par



We introduce the Capsule Summarization (CapsSum) layer and summarize the generated capsules into only a few. The reduction in the number of capsules reduces the number of parameters and speeds up the network.


\par
Deeper networks deeper usually have higher representation ability. Therefore, we started by making the network deeper. Stacking several fully-connected capsule layers with DR inferring the capsules from one layer to the next is computationally expensive.
Moreover, using the DR algorithm multiple times, leads to poor learning in the intermediate layers \cite{Xi2017}. However, there are different DR alternatives, and as we show, employing a three-dimensional 3DR algorithm \cite{Rajasegaran}) is a reasonable alternative. 
\par

We introduce Multi-Level Capsule Extractor (MLCE), that uses the 3DR algorithm twice and includes two CapsSum layers. MLCE takes capsules as input, and outputs a combination of high-level and low-level capsules. MLCE consists of convolutional layers,  CapsSum, and 3DR. It generates fewer capsules in two levels that provide a robust part-to-whole representation and reduces the total number of parameters in the network. 

\par

Networks based on CapsNet usually include a decoder to reconstruct the input images and avoid over-fitting. In order to maintain accuracy, we carefully employ an efficient decoder (class-independent decoder). 

\par 
In summary, we propose DL-CapsNet as a deep, light and highly accurate variant of CapsNet by using the MLCE module on top of a deep convolutional sub-network, and employing an efficient decoder.
Our contributions are:

\par

\par


\begin{itemize}
    \item We develop a deep network achieving high test accuracy. DL-CapsNet includes several capsule layers, and uses 3DR twice to make a high-level representation of the input images. We achieve 91.23\% accuracy for CIFAR-10 using a 7-ensemble model.
    
    \item We reduce  complexity by introducing the CapsSum layer. CapsSum reduces the number of generated capsules by using a deep structure consisting of several capsule layers. DL-CapsNet contains 6.8M parameters.
    
    \item We evaluated the network for CIFAR-10, SVHN, and Fashion-MNIST and achieved state-of-the-art results. In addition, we support more complex datasets. Using a 7-ensemble model for CIFAR-100, we achieve 68.36\% accuracy.

    
    
    
\end{itemize}
\par
The rest of this paper is organized as follows. The related works are presented in Section 2. Section 3 reports the background. DL-CapsNet is presented in Section 4. Experiments and results are presented in Section 5. The paper is concluded in Section 6.

\section{Related Works}
Several studies have improved the CapsNet networks. Yang et al. \cite{Yang2020} proposed RS-CapsNet. RS-CapsNet integrates Res2Net blocks to extract features in multiple scales. It also uses the Squeeze-and-Excitation (SE) block to emphasize more useful features. In order to enhance the representation ability and reduce the number of capsules, RS-CapsNet uses a linear combination 
of capsules. 

\par
Huang et al. proposed DA-CapsNet \cite{Huang2020}, which uses the attention mechanism after the convolution layers and also after the primary capsules layer.
DA-CapsNet is highly accurate for SVHN, CIFAR10, FashionMNIST, smallNORB, and COIL-20 datasets. DA-CapsNet outperforms CapsNet in image reconstruction.



Shiri et al. \cite{Shiri2020} proposed Quick-CapsNet (QCN) modifying the low-level feature extraction, resulting in only a few capsules. The reduction of PCs results in a significant speedup at the cost of marginal loss of accuracy. QCN uses an optimized decoder to improve the network generalization. 

\par
Shiri et al. \cite{Shiri2021} proposed CFC-CapsNet. CFC-CapsNet used a new layer for creating PCs out of the extracted features. This layer results in fewer number of capsules while improving accuracy. The reduction of number of capsules, reduces the overall number of parameters as well as speeding up the network.

\par
Deliege et al. \cite{Deli2018} proposed HitNet which replaces a layer with a Hit-or-Miss layer. Capsules in this layer are trained to hit or miss a central capsule. To this end, a specific loss function is used. This network  contains a reconstruction sub-network synthesizing samples of images. The reconstruction sub-network can be used as an augmentation method to avoid overfitting. HitNet uses ghost capsules to detect mislabeled data in the training set.

\par
He et al. \cite{He2019} proposed Complex-Values Dense CapsNet (Cv-CapsNet) and Complex-Valued Diverse CapsNet (Cv-CapsNet++). Both networks include a complex-valued sub-network for extracting features in different scales. Afterwards, complex-valued PCs are created out of the extracted features. Cv-CapsNet++ implements a hierarchy of three-level Cv-CapsNet model and hence produces multi-dimensional complex valued PCs.

\par

Chen et al. \cite{chen2020} propose a deep capsule network combined with a U-Net preprocessing module (DCN-UN) which attempts to improve CapsNet for complex datasets such as CIFAR-10 and CIFAR-100. A convolutional capsule layer is developed based on local connections and weight-sharing strategies which allows reducing the number of parameters. DCN-UN employs a greedy strategy to develop the Mask Dynamic Routing (MDR) to improve the performance.

\par

Ayidzoe et al. \cite{AbraAyidzoe2021} introduce a less complex variant of CapsNet with an improved feature extractor. They employ a Gabor filter and customized blocks for preprocessing data leading to the extraction of the semantic information. This results in enhanced activation diagrams and learns the hierarchical information meaningfully.

\par

Tao et al. \cite{Tao2022} present an efficient and flexible network based on capsules referred to as Adaptive Capsule CapsNet (AC-CapsNet). This network replaces the primary capsules with an adaptive capsule layer. The adaptive values contain both the spatial information of each capsule vector and the local relationship among the neurons contained in the capsules.

\par
This work takes a similar approach as the works explained above. However, we take into account the network accuracy and size (number of parameters) simultaneously. As we later show, we achieve the highest accuracy among the state-of-the-art networks based on capsule. 

\section{Background}

In this Section, we review background. We explain normal and 3D CapsCells, the normal and the 3D routing algorithms, and the class-independent decoder. These units along with the CapsSum and  MLCE (explained in Section 4) build DL-CapsNet.

\subsection{Capsule Cell}

\begin{figure}[htp]
    \centering
    \includegraphics[keepaspectratio, scale=0.3]{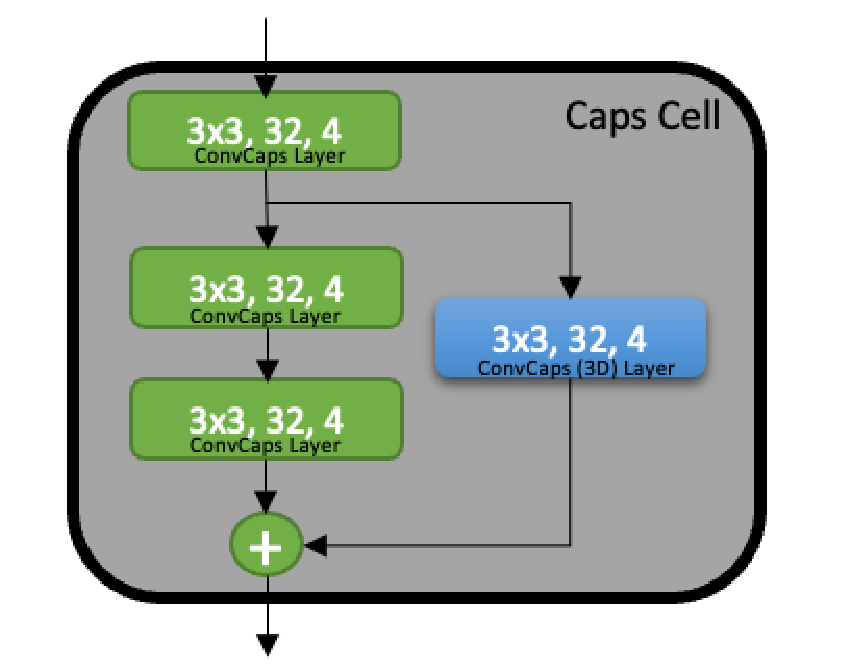}
    \caption{The architecture of a CapsCell with $K=3$, $D=4$ and $N_v=32$. This unit contains several ConvCaps layers and a skip-connection. For the 3DR CapsCells, the skip connection performs the 3D dynamic routing operation. } 
    \label{fig:capscell}
\end{figure}

Capsule Cells (CapsCells) were introduced by Rajasegaran et al. \cite{Rajasegaran2019} as units including a combination of several Convolutional Capsule (ConvCaps) layers and a skip connection. The ConvCaps layer is a convolutional layer with its outputs reshaped to capsules. Figure \ref{fig:capscell} shows a CapsCell, which includes three ConvCaps layers. The output of the first layer is skip connected to the output of the last layer. This is done to avoid the problem of vanishing gradients. In addition, the skip-connection helps route the low-level capsules to high-level capsules. The skip-connection either includes a ConvCaps layer, or implements the 3D dynamic routing (3DR) algorithm. The former is called a normal CapsCell and the latter is called a 3DR CapsCell.

\par

Each ConvCaps layer has 3 parameters: $K$ or the kernel size, $D$ or the number of values for each output vector (the dimensionality), and $N_v$ or the number of vectors per spatial location of the output feature map. 


\begin{figure*}[htpb]
    \centering
    \includegraphics[width=\textwidth,keepaspectratio]{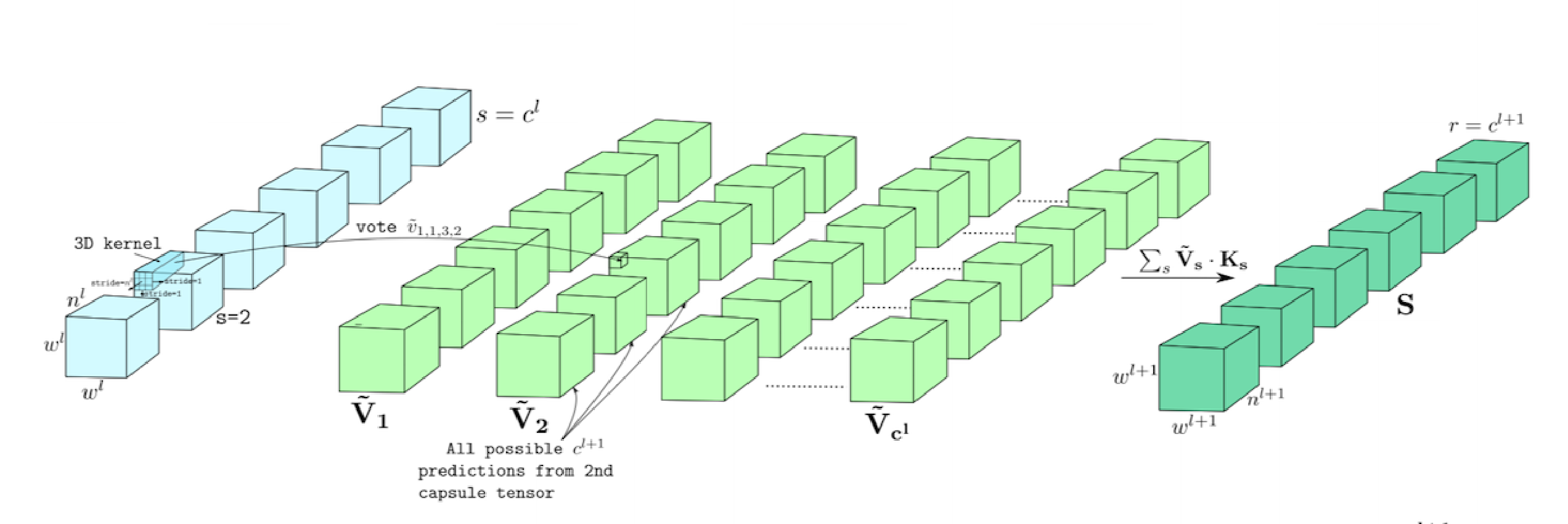}
    \caption{3D-Routing method. Each capsule in layer $l$, predicts $c^{l+1}$ capsules. As a result, there are $c^l$ predictions for a capsule in layer $l+1$. \cite{Rajasegaran} }
    \label{fig:deepcaps-3dr}
\end{figure*}

\subsection{Routing Capsules}
In this Section we explain two routing capsule methods in subsequent layers: Dynamic Routing (DR) and three-dimensional Dynamic Routing (3DR). 

\subsubsection{Dynamic Routing}
In this method, all input capsules contribute to forming any of the output capsules.  DR finds a coefficient for each input capsule and works as a routing method to relate the input capsules to the output capsules. The coefficients are not trained. Instead, DR creates the output capsules iteratively during training based on the agreement between the input capsules.
\par
Capsules in the lower-level (input capsules) need to decide how to send their vector to the output capsules (higher-level capsules). This decision is made by changing a scalar implying the weight of the capsule. This scalar is multiplied by the vector and fed as input to the output capsules. The output capsules are a weighted sum of the input capsules, with the weights determined by DR.


\subsubsection{3D Dynamic Routing}
DR routes capsules in a global manner, since all input capsules contribute to all output capsules. The 3DR algorithm, performs the routing locally. Capsules coming from nearby regions of the previous feature map
are routed together to output capsules. Figure \ref{fig:deepcaps-3dr} depicts how 3DR groups the input capsules and routes them to the output capsules. A capsule in layer $l$, predicts a $c^{l+1}$ number of capsules. Therefore, for each capsule in layer $l+1$, there are $c^l$ predictions. $s$ and $S$ denote the input and output capsules respectively, and $\hat{V}$ are the intermediate votes in the routing algorithm. 
Like DR, the weights are iteratively inferred, and not trained.


\subsection{Class-Independent Decoder}
CapsNet comes with a basic decoder based on Fully-Connected (FC) layers. The output capsules are fed to this decoder to reconstruct the input images. To regularize the training process and avoid overfitting, the reconstructed images are compared to input images. The result is considered inside the loss function (reconstruction loss). We use the class-independent decoder introduced by Rajasegaran et al. \cite{Rajasegaran}.

This decoder comes with two important benefits. First, it is based on deconvolution. Deconvolutional layers capture spatial relationships better than FC layers and include fewer number of parameters. Second, the decoder drops the incorrect capsules and removes them from the reconstruction process which leads to a more robust reconstruction. 
Sabour et al. \cite{Sabour2017} masked the incorrect capsules with zeros. The class-independent decoder, discards the incorrect capsules completely. As for all different categories (classes) there is a fixed vector of data kept and used for reconstruction, this decoder is class-independent as all classes are treated similarly. Experiments show that class-independence makes the decoder more robust \cite{Rajasegaran2019}. 

\section{DL-CapsNet}

\begin{figure*}[htpb]
    \centering
    \includegraphics[width=0.7\textwidth,keepaspectratio]{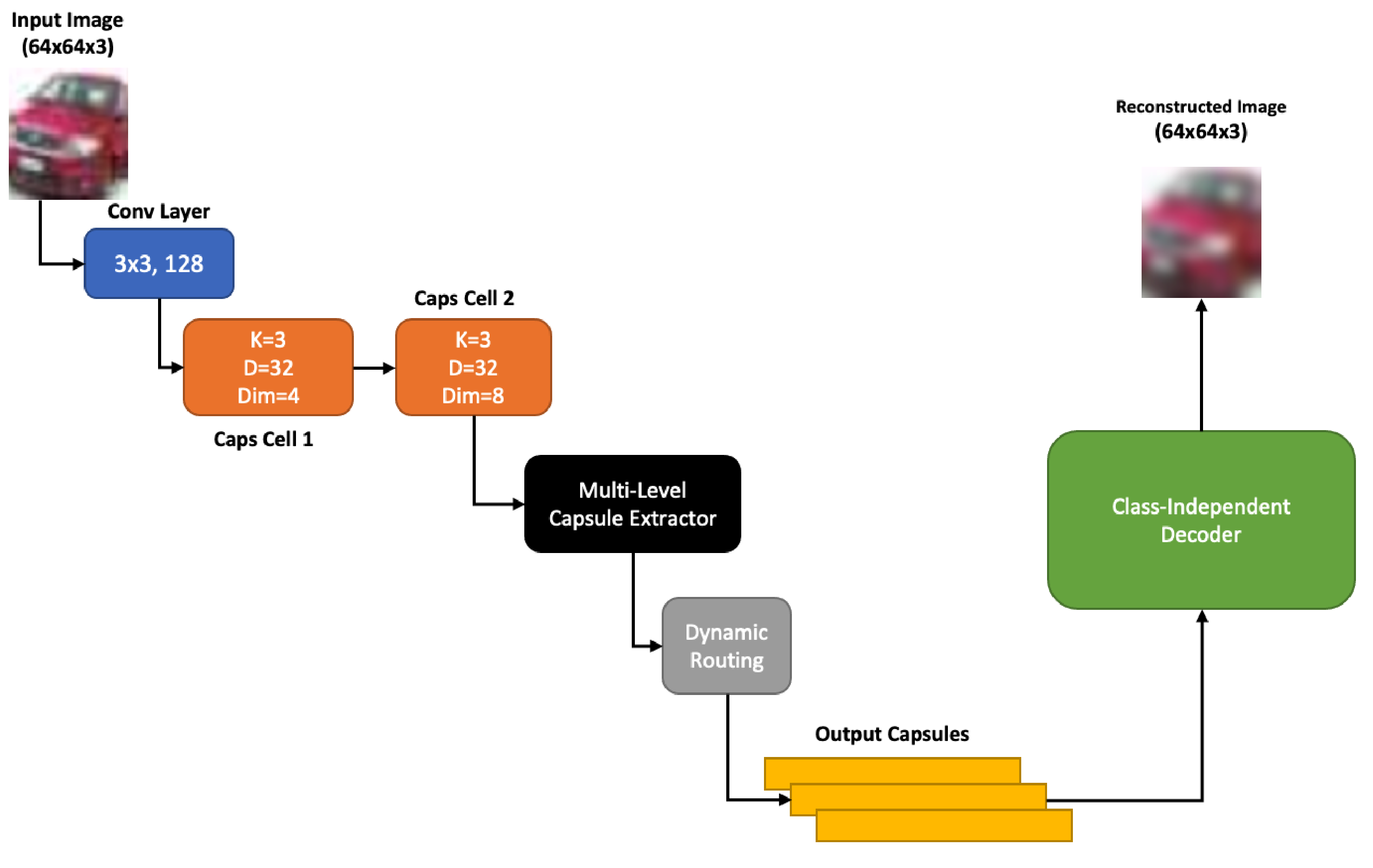}
    \caption{DL-CapsNet architecture. The network includes two CapsCells, the MLCE module.}
    \label{fig:dlcaps-arch}
\end{figure*}

Figure \ref{fig:dlcaps-arch} shows the architecture of DL-CapsNet. The network consists of a convolutional layer, two normal CapsCells, MLCE, the DR section and the class-independent decoder. 
\par 
DL-CapsNet uses a convolutional layer to extract very low-level features. Features are reshaped to capsules. Afterwards, there are two normal CapsCells to create richer capsules. The output of the second CapsCell is fed to the MLCE module. MLCE is presented in the next Section. The capsules generated by MLCE are used to infer the output capsules of the network using DR.  Similar to CapsNet, classification is done based on the output capsules. There are $K$ capsules where $K$ is the number of classes in the classification task, and the capsule with the highest length (L2 Norm) corresponds to the predicted class. These capsules are also fed to the decoder network.

\subsection{Capsule Summarization (CapsSum) layer}
\begin{figure}[H]
    \centering
    \includegraphics[keepaspectratio,scale=0.2]{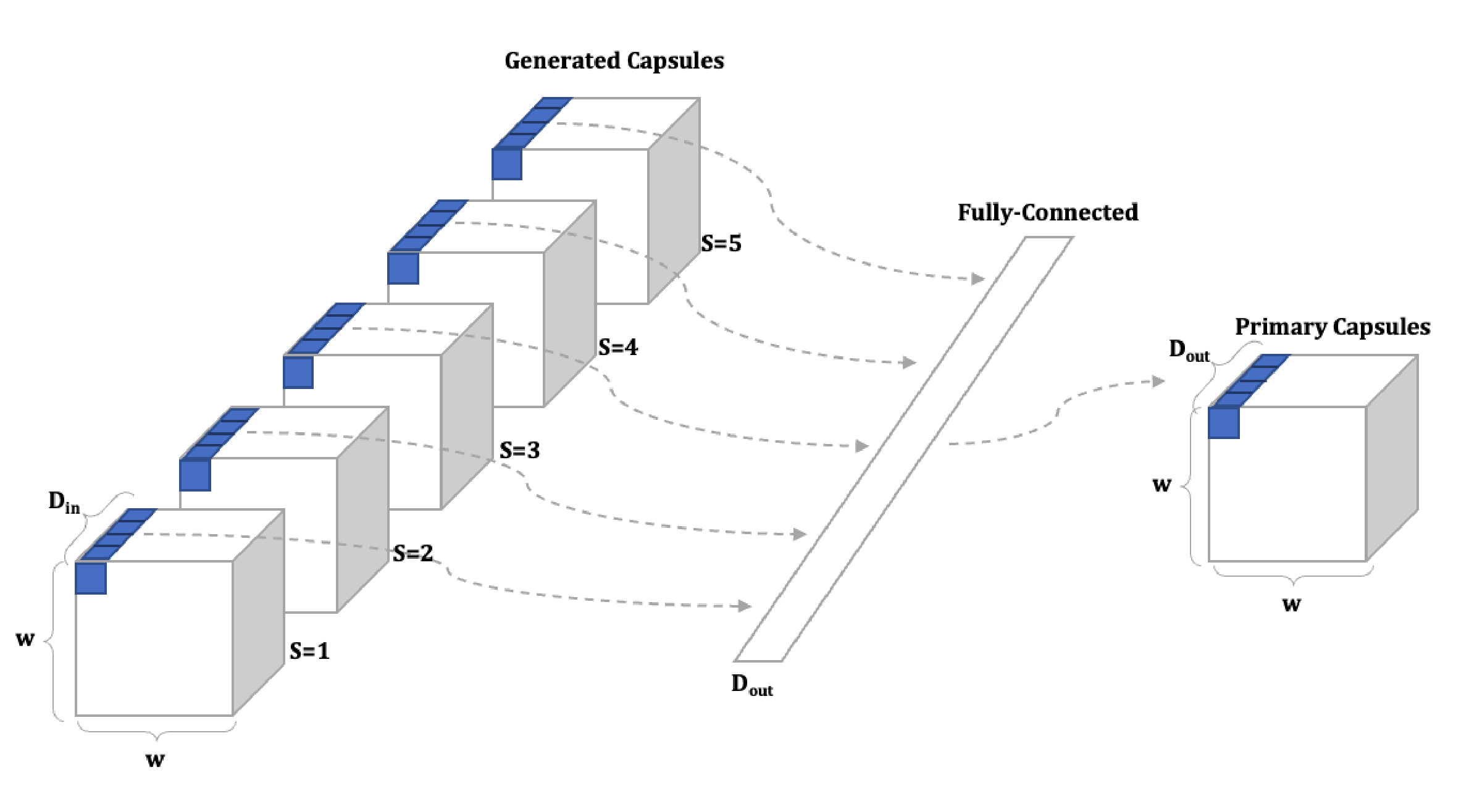}
    \caption{The capsule summarization layer. A total of $w \times w \times 5$ generated capsules are summarized into $w \times w \times D_{out}$ primary capsules using $w \times w$ Fully-Connected (FC) layers. The first FC layer is shown.}
    \label{fig:capssum}
\end{figure}

We propose the CapsSum layer to reduce the number of generated capsules. CapsSum is inspired by the Convolutional Fully-Connected CapsNet(CFC)  \cite{Shiri2021}. 
 As stated earlier, CapsNet includes a simple feature extractor 
including two convolutional layers. Originally, the extracted data is directly reshaped to capsules. Alternatively, the CFC layer was proposed for translating the low-level extracted features to fewer capsules, resulting in parameter reduction and network speed-up. We customize and integrate the CFC layer in DL-CapsNet. We summarize the generated capsules by the DeepCaps network to produce a new set of PCs. To this end, we introduce a capsule summarization layer. Figure \ref{fig:capssum} shows how this layer works. The nearby generated capsules 
are all flattened, and fed to a fully-connected layer to produce a single capsule. This procedure is repeated for all spatial locations in the generated capsules. There are a total of the $S\times w \times w$ generated capsules each with $D_{in}$ elements. For each spatial location $(i,j)$ where $i,j \in [1,w]$, a total of $S$ capsules with $D_{in}$ dimensionality are collected, flattened and fed to a single fully-connected layer to produce a single capsule with a different dimensionality $D_{out}$. It is noteworthy that the figure shows the procedure for the first spatial location (1,1). There are $w \times w$ fully-connected layers to summarize the entire generated capsules to the PCs. The proposed layer, reduces the number of capsules $S$ times. Intuitively, each output capsule corresponds to a set of nearby capsules. We reduce all those correlated nearby capsules to a single capsule using a fully-connected layer. The reduction process includes trainable parameters (contained in the fully-connected layer). However, the reduction in the number of capsules outnumbers the increase in the parameters.

\subsection{Multi-Level Capsule Extractor (MLCE) module}
MLCE has two goals. First, to create a rich and robust representation of the input images using the extracted capsules, and second, reduce network size
(in terms of the number of parameters).

\par



To meet the first goal, we stack two 3DR CapsCells. 3DR is not as computationally expensive as DR  due to performing the routing in a localized manner. As a result, it is possible to stack two 3DR CapsCells to make a deep representation of data.
\par

To meet the second goal, we use CapsSum.  The primary capsules are multiplied by weight matrices each of which corresponds to one of the categories in the classification task. Therefore, reducing the number of PCs results in fewer weight matrices and as a result fewer overall number of parameters. In addition, the more primary capsules are in the network, the more computationally expensive DR would become. The more primary capsules, the more time it takes for the DR algorithm to infer the output capsules from input capsules
We use two instances of the CapsSum layer inside MLCE to reduce the number of generated capsules number of PCs.
\par

\begin{figure}[htp]
    \centering
    \includegraphics[keepaspectratio, scale=0.22]{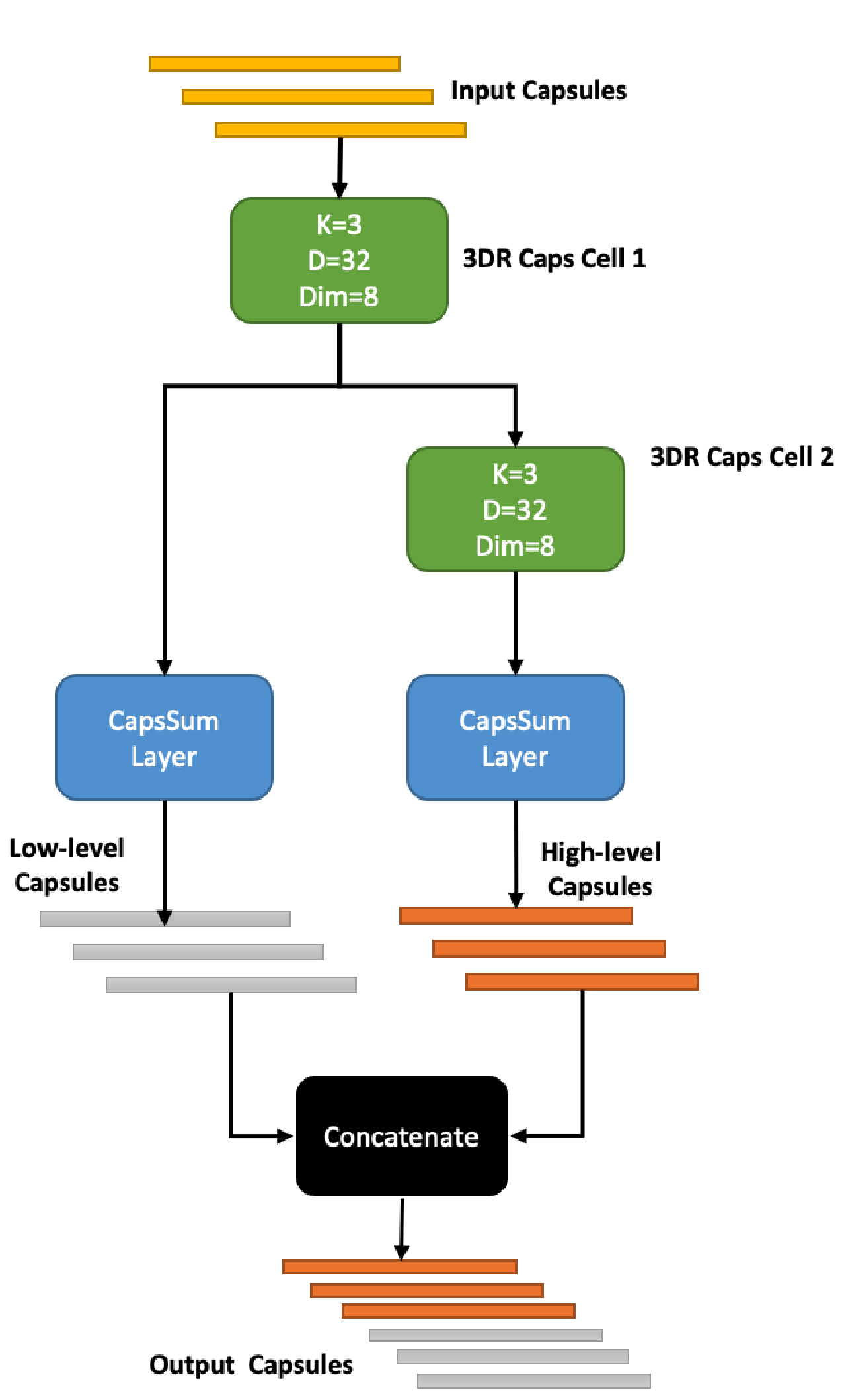}
    \caption{The architecture of MLCE module. The module consists of two 3DR CapsCells and two CapsSum layers.}
    \label{fig:mlce-arch}
\end{figure}

MLCE is depicted in Figure \ref{fig:mlce-arch}. MLCE infers two sets (levels) of capsules from the input capsules, and concatenates them to generate a combination of low-level and high-level capsules.
This module stacks two 3DR CapsCells. First, the input capsules are fed to a 3DR CapsCell. The first set of output capsules (low-level capsules) are formed by using a CapsSum layer on top of this 3DR CapsCell. Afterwards, there is another 3DR CapsCell, and another CapsSum layer is used to summarize the high-level capsules into a few capsules (high-level capsules).
\par
The 3DR CapsCells create a robust part-to-whole representation of data by localizing the routing process. Stacking two of these layers, results in creating a deep representation of data. In addition, combining the output of the first and second CapsCells ensures that the created capsules include both low-level and high-level information which results in an increased generalization ability of the network. On the other hand, the CapsSum layer 
improves the representation, reduces the number of parameters, and enhances network speed.
We use a CapsSum layer after each 3DR CapsCell. The combination of 3DR routing and the CapsSum layer in two levels, enables the network to form a multi-level representation of the input image.

\par

\subsection{Loss function}
Our loss function is similar to the one introduced by Sabour et al. \cite{Sabour2017} (margin loss). This loss function considers penalties for incorrect predictions and disregards predictions with a very high or very low probability:
\[ L_k = T_k \max(0,m^+-||V_k||)^2 + \lambda(1-T_k)\max(0,||V_k||-m^-)^2 \]
In this equation, $L_k$ is the loss term for capsule $k$, $T_k$ is 0 for incorrect class and 1 otherwise, $m^+$ and $m^-$ are used to disregard high or low probabilities, and lambda is used for controlling the gradient at the start of the training.

\section{Experiments and Results}
In this Section, we explain the experiments and the corresponding results.

\subsection{Datasets}

We test DL-CapsNet for datasets commonly used for testing CapsNet and its variants: Fashion-MNIST (FMNIST), SVHN, CIFAR-10 and CIFAR-100. Testing CIFAR-100 is possible due to the small number of capsules generated by the MLCE module. For SVHN, CIFAR-100 and CIFAR-10 datasets, the input images are resized from $32\times32\times3$ to $64\times64\times3$ and for F-MNIST the original images are used throughout the experiment. We do the resizing because the images in these datasets include richer features compared to the F-MNIST dataset.

\subsection{Experiment Settings}
We modify and use several units introduced in DeepCaps \cite{Rajasegaran}. We implement DL-CapsNet using the Keras implementation of DeepCaps \footnote{https://github.com/brjathu/deepcaps}. We use a NVIDIA 2080Ti GPU for running the experiments. Following the DeepCaps implementation, we perform hard training in all the experiments. After training a network for the first 100 epochs, we tighten the bounds of the loss function by changing the values for $m^+$ and $m^-$ and train the network for another 100 epochs. The experiments are repeated 5 times. We report the average values due to the little variation in the results. We used Adam optimizer with starting learning rate of 0.001. We also use an exponential decay ($\gamma=0.96$) and batch size of 128.



\subsection{Network Accuracy}

Table \ref{table:tres_acc} compares the network classification accuracy of DL-CapsNet to some state-of-the-art CNN networks (shown in the top part of the Table), and some recent and efficient variants of CapsNet (shown in the middle of the Table). DL-CapsNet and other recent CapsNet variants fall behind powerful CNN networks such as BiT-M \cite{Kolesnikov2019}. As the Table shows, DL-CapsNet achieves competitive accuracy for all datasets compared to the state-of-the-art. To further improve the classification accuracy, we use a 7-ensemble model of DL-CapsNet. In this method, seven instances of DL-CapsNet are trained and the softmax outputs are averaged to determine the final output of the network. Using this method, DL-CapsNet reaches 91.29\% and 68.36\% accuracy for CIFAR-10 and CIFAR-100 datasets.
\par
 For datasets with a high number of classes such as CIFAR-100, the number of parameters can be very high. Therefore, the DR algorithm can take a long time to infer the output capsules.
The 7-ensemble model of DL-CapsNet obtains 68.36\% accuracy for the CIFAR-100 dataset. 
 For the rest of the datasets i.e. CIFAR-10, SVHN and Fashion-MNIST, DL-CapsNet is among the powerful CapsNet-based networks in terms of the accuracy (DeepCaps \cite{Rajasegaran2019} and RS-CapsNet \cite{Yang2020}.

\newcommand{\specialcell}[2][c]{%
  \begin{tabular}[#1]{@{}c@{}}#2\end{tabular}}
  
\begin{table}[htpb]
\caption{Classification accuracy of some state-of-the art CNNs (shown on top) and the state-of-the-art CapsNet variants (shown in the middle) compared to DL-CapsNet (shown on the bottom). We obtain competitive accuracy on all datasets.} 
\centering 
\resizebox{\linewidth}{!}{
\begin{tabular}[width=\linewidth]{|l|c| c| c|c|}
 \hline
\textbf{Model} & \textbf{CIFAR-100} & \textbf{CIFAR-10} & \textbf{SVHN} & \textbf{FMNIST} \\ [0.5ex] 
\hline 

DenseNet \cite{Huang2016} & 82.4\%/15.3M & 96.40\%/15.3M & 98.41\%/15.3M & 95.40\%/15.3M  \\ [1ex] 
RS-CNN \cite{Yang2020} & 61.14\%/2.8M & 90.15\%/2.7M & 95.56\%/2.7M & 93.34\%/2.7M \\ [1ex]
BiT-M \cite{Kolesnikov2019} & 92.17\%/928M & 98.91\%/928M & - & - \\ [1ex]
\hline
DA-CapsNet \cite{Huang2020} & - & 85.47\%/- & 94.82\%/- & 93.98\%/- \\ [1ex]

CapsNet (7-ens) \cite{Sabour2017} & - & 89.40\%/11.7M & 95.70\%/11.7M & - \\ [1ex] 
Cv-CapsNet++ \cite{He2019} & - & 86.70\%/2.7M & - & 94.40\%/2.5M \\ [1ex]
CFC-CapsNet \cite{Shiri2021}  & - & 73.15\%/5.9M & 93.29\%/5.9M & 92.86\%/5.7M \\ [1ex]
HitNet \cite{Deli2018} & - & 73.30\%/8.9M & 94.50\%/8.9M & 92.30\%/8.9M \\ [1ex]
RS-CapsNet \cite{Yang2020} & 64.14\%/16.8M & 91.32\%/5M & 97.08\%/5M & 94.08\%/5M \\ [1ex]
DCN-UN MDR \cite{chen2020} & 60.56\%/4.8M & 90.42\%/1.4M & - & 93.33\%/- \\ [1ex]
Gabor CapsNet \cite{AbraAyidzoe2021} & 68.17\%/22.6M & 85.24\%/10.4M & - & 94.78\%/- \\[1ex]
AC-CapsNet \cite{Tao2022} & 66.09\%/4.12M & 92.02\%/1.26M & 96.86\%/1.26M & - \\[1ex]
DeepCaps (7-ens) \cite{Rajasegaran2019} & - & 92.74\%/13.4M & 97.56\%/13.4M & 94.73\%8.5M \\ [1ex]
\hline
DL-CapsNet & 63.73\%/11.2M & 89.06\%/6.8M & 95.82\%/6.8M & 94.21\%/6.4M \\ [1ex]
DL-CapsNet (7-ens) & 68.36\%/11.2M & 91.29\%/6.8M & 97.09\%/6.8M & 94.72\%/6.4M \\

\hline 
\end{tabular}}
\label{table:tres_acc} 
\end{table}

\subsection{Number of Parameters}

Table \ref{table:tres_acc} shows the number of parameters besides the network inference accuracy for each dataset.
Some recent and powerful CapsNet and CNN variants employ novel solutions for reducing the number of weights and include a significantly fewer number of parameters. These networks however, obtain a lower accuracy compared to DL-CapsNet. For example, DCN-UN MDR and AC-CapsNet include 4.8M and 4.12M parameters for the CIFAR-100 dataset (compared to 11.2M in DL-CapsNet), however our proposed network achieves a slightly higher accuracy. In addition, in contrast to other works, we also report the network speed by showing how our network performs in terms of the network inference time.


\subsection{Network Training and Inference Time}
There are only few works in the CapsNet domain reporting the inference time. In addition, providing a fair comparison requires using the same base implementation for CapsNet, and the same GPU. Only the DeepCaps \cite{Rajasegaran} network satisfied these two condition. 
Using a Geforce 2080Ti GPU, the inference in DL-CapsNet takes 2.18ms for a single $64\times64\times3$ image of the CIFAR-10 dataset. This is 1.97x faster than DeepCaps, as it takes 4.30ms for DeepCaps to do the same job.



\section{Conclusion}
We present DL-CapsNet as an efficient and effective CapsNet variant. Despite the deep structure of the network, DL-CapsNet is still a light, yet highly accurate network. Using a 7-ensemble model, DL-CapsNet achieves a competitive accuracy of 91.29\% for the CIFAR-10 dataset using 6.79M parameters. With 68.36\% test accuracy for CIFAR-100, DL-CapsNet performs well on complex datasets with a high number of categories.

\section*{Acknowledgment}
This research has been funded in part or completely by the Computing Hardware for Emerging Intelligent Sensory Applications (COHESA) project. COHESA is financed under the National Sciences and Engineering Research Council of Canada (NSERC) Strategic Networks grant number NETGP485577-15.

\bibliographystyle{unsrt}

\end{document}